%% file: main.tex
\documentclass{article}

\usepackage[dblblindworkshop,final]{neurips_2026_vericode}
\workshoptitle{AI for Verifiable Coding (VeriCodeGen)}

\usepackage[utf8]{inputenc}
\usepackage[T1]{fontenc}
\usepackage{hyperref}
\usepackage{url}
\usepackage{booktabs}
\usepackage{amsmath}
\usepackage{amssymb}
\usepackage{microtype}
\usepackage{xcolor}
\usepackage{graphicx}
\usepackage{multirow}
\usepackage{array}
\usepackage{listings}
\usepackage{placeins}
\title{How Much Can Repeated Running Buy? A Measured Variance Ceiling for LLM Code Evaluation\thanks{This paper is a substantially revised and retitled version of arXiv:2606.00920v1 (``Accuracy, Stability, and Repeated-Run Reliability of Large Language Models on Deterministic Programming Tasks''). Both versions analyse the same 16-model, 99-problem, five-run outcome data; this version reframes the analysis as a variance decomposition. Problem list, binary outcome tensor, and analysis code: \url{https://github.com/Yongxi-Zhou/leetcode-evaluator-data-output}.}}

\author{%
  Yongxi Zhou \\
  Northeastern University \\
  \texttt{zhou.yongx@northeastern.edu} \\
  \And
  Lai Yun Choi \\
  Northeastern University \\
  \texttt{choi.lai@northeastern.edu} \\
  \And
  Jiaxi Wen\thanks{Corresponding author.} \\
  Northeastern University \\
  \texttt{wen.jiax@northeastern.edu} \\
  \And
  Wenbo Ye \\
  University of Southern California \\
  \texttt{ywb6113@gmail.com} \\
  \And
  Junwei Yao \\
  Northeastern University \\
  \texttt{yao.junw@northeastern.edu} \\
}

\begin{document}

\maketitle

\begin{abstract}
A benchmark score is a measurement whose uncertainty has two sources: an aleatoric component from stochastic decoding and an epistemic component from the finite problem set. Using 16 models on 99 recent LeetCode-style problems with two prompts and five runs each (15{,}840 judged runs in a fully crossed design), we estimate both components and ask a design question: how much can repeated running buy? First, the pooled-binomial interval that repeated-run protocols invite answers neither question of interest---it is about $2\times$ too narrow for generalisation to new problems and $2\times$ too wide for the fixed problem set---and under correctly clustered intervals, 12 of 15 adjacent leaderboard pairs are statistically unresolved. Second, the \emph{variance ceiling} $1+\mu_w/\sigma_b^2$, the largest factor by which repetition can reduce score variance, has median 1.34 across models (range 1.05--1.71), so repetition can narrow confidence intervals by at most 23\% on this benchmark, and by 14\% for the median model. Third, once the cost of adding verified problems is priced in, the precision-optimal repeat count is $R^{*}=\sqrt{\smash[b]{\mu_w c_p/(\sigma_b^2 c_g)}}$, and the variance curve is flat around it: on this benchmark the decisive reason to repeat is not precision but \emph{identification}---two to three runs suffice to estimate the components that the ceiling and the fixed-benchmark interval require. All three are measurements of one benchmark at one decoding temperature, and we state explicitly what they do and do not license.
\end{abstract}

\section{Introduction}
Because decoding is stochastic, the same model can pass a coding problem on one invocation and fail on the next, and a growing number of evaluation protocols respond by running each problem several times \citep{brown2024large,atil2024llm,alvarado2025repetitions}. The practice is usually motivated as making the reported score more trustworthy. Whether it does depends on an accounting that is rarely performed: the sampling variance of a benchmark score has an \emph{aleatoric} part contributed by decoding, which repetition averages away, and an \emph{epistemic} part contributed by the finite problem sample, which repetition cannot touch \citep{hullermeier2021aleatoric,baan2023uncertainty}.

This accounting matters most where verification is strongest. When the verifier is deterministic and machine-checkable---executable test suites here, proof obligations or model checkers in verified code generation more broadly---the verdict on any given program is exact (up to the timing noise of resource-limit verdicts), so the uncertainty of a reported benchmark score comes almost entirely from the two sources above: which problems the benchmark contains, and which programs the model happens to sample. Benchmarks and metrics for verifiable-coding pipelines inherit both components, and trusting the guarantees they advertise requires knowing how much of the score's variance each source contributes.

This paper performs that accounting on one deterministic code-generation benchmark and reports three measurements aimed at evaluation design.
\begin{enumerate}\itemsep1pt
  \item \textbf{Leaderboards collapse under correct intervals.} Pooling $N\times R$ runs into one binomial answers neither estimand: the resulting interval is a median $2.05\times$ too narrow if the target is generalisation to new problems and $1.96\times$ too wide if the target is the fixed benchmark. With clustered intervals, 12 of 15 adjacent pairs in our 16-model leaderboard cannot be separated at the 95\% level (\S\ref{sec:collapse}).
  \item \textbf{A measured variance ceiling.} The factor $1+\mu_w/\sigma_b^2$ bounds how much score variance repetition can remove. Its per-model values span 1.05--1.71 (median 1.34), i.e.\ confidence intervals can shrink by at most 2--23\% per model, 14\% for the median model; the bound is stable across our two prompt conditions (\S\ref{sec:ceiling}).
  \item \textbf{Cost-aware allocation.} Under a generation-only budget, one run per problem minimises variance; but verified problems, not generations, are the expensive asset. Two-stage sampling gives $R^{*}=\sqrt{\smash[b]{\mu_w c_p/(\sigma_b^2 c_g)}}$, and the flatness of the variance curve implies that, on this benchmark, the real reason to repeat is identification of the components themselves (\S\ref{sec:alloc}).
\end{enumerate}
All three are measurements on one benchmark at one temperature; \S\ref{sec:scope} states what they do and do not support.

\section{Position Among Prior Work}
The decomposition we use is classical---it is the one-facet crossed design of generalizability theory \citep{brennan2001generalizability} and the textbook variance-components model---and recent work has begun applying exactly this machinery to LLM evaluation. \citet{miller2024adding} derives clustered standard errors and resampling guidance for evals; \citet{bowyer2025dont} shows CLT-based intervals fail on small benchmarks; \citet{luettgau2025hibayes} fits hierarchical Bayesian GLMs to evaluation data; \citet{hariri2025dont} replaces pass@$k$ with posterior estimates of success probability. Our contribution is therefore not the statistics but the \emph{measurements}: to our knowledge, the first per-model estimates on a code benchmark of (i) the aleatoric/epistemic split of score variance, (ii) the resulting ceiling on what repetition can buy, with uncertainty on the ceiling itself, and (iii) the empirical intraclass correlations (0.59--0.95) that let practitioners predict design effects for protocols like ours, at this temperature, without rerunning anything. We also connect the split to evaluation economics ($R^{*}$), which the works above do not address. Complementary evidence that repeated scores vary comes from \citet{atil2024llm,alvarado2025repetitions}; concerns about benchmark statistical power are long-standing \citep{card2020little,dror2018hitchhiker,bowman2021will}. Unlike output-level uncertainty quantification \citep{kuhn2023semantic,guo2017calibration}, we quantify uncertainty of the \emph{score}.

\section{Two Sources of Uncertainty}
\label{sec:sources}
Let $Y_{i,r}\in\{0,1\}$ indicate whether run $r\in\{1,\dots,R\}$ on problem $i\in\{1,\dots,N\}$ passes a deterministic verifier, $\hat p_i=\frac1R\sum_r Y_{i,r}$, and let the reported score be $\mathrm{RLPR}=\frac1N\sum_i \hat p_i$. Model each problem as carrying a latent success probability $p_i$ with population mean $\mu$ and variance $\sigma_b^2$, and runs as conditionally independent Bernoulli draws. Then
\begin{equation}
\mathrm{Var}(\mathrm{RLPR})=\frac{1}{N}\Big(\underbrace{\sigma_b^2}_{\text{epistemic}}+\underbrace{\mu_w/R}_{\text{aleatoric}}\Big),
\label{eq:var}
\end{equation}
with $\mu_w=\mathbb{E}[p_i(1-p_i)]$. We estimate $\mu_w$ by the unbiased $\mathrm{AV}=\frac1N\sum_i\frac{R}{R-1}\hat p_i(1-\hat p_i)$ and $\sigma_b^2$ by moment matching, $\widehat{\sigma_b^2}=s^2(\hat p)-\mathrm{AV}/R$. The moment estimator can go negative when $\sigma_b^2\!\approx\!0$; in our data it never does (16/16 positive). Three derived quantities organise the paper:

\paragraph{ICC.} $\rho=\sigma_b^2/(\sigma_b^2+\mu_w)$ is the intraclass correlation of runs within a problem. It predicts the \emph{design effect} of pooling: treating $NR$ clustered runs as independent understates variance by $1+(R-1)\rho$, i.e.\ intervals are $\sqrt{1+(R-1)\rho}$ too narrow.

\paragraph{Ceiling.} Relative to a single run, infinite repetition reduces the variance in Eq.~\ref{eq:var} by at most
\begin{equation}
\frac{\mathrm{Var}(R{=}1)}{\mathrm{Var}(R{=}\infty)}=1+\frac{\mu_w}{\sigma_b^2},
\label{eq:ceiling}
\end{equation}
a benchmark- and model-specific quantity that is estimable from a repeated-run pilot and requires no cost assumptions.

\paragraph{Optimal $R$.} With per-problem acquisition cost $c_p$ and per-generation cost $c_g$, minimising Eq.~\ref{eq:var} under budget $B=N(c_p+Rc_g)$ gives $R^{*}=\sqrt{\smash[b]{\mu_w c_p/(\sigma_b^2 c_g)}}$. A generation-only budget ($c_p{=}0$) recovers the $R{=}1$ rule; costly problems push $R^{*}$ up.

\paragraph{Estimand caveat.} Interpreting $\sigma_b^2$ as epistemic sampling error treats the $N$ problems as draws from a task population. Under the alternative reading---the benchmark itself is the estimand---$\sigma_b^2$ is descriptive dispersion, and the only sampling uncertainty is aleatoric, $\mathrm{Var}=\mu_w/(NR)$. We report results under both readings.

\section{Setup}
\label{sec:setup}
\paragraph{Benchmark.} We use the $N{=}100$ most recent public, non-premium LeetCode problems at collection time (collected 2026-03-22; problem numbers 3759--3873; 18 Easy / 55 Medium / 27 Hard). Selection is by recency rank alone, which limits but does not rule out contamination (\S\ref{sec:scope}). Every generation is a Python~3 solution submitted to LeetCode's official judge; a run is scored 1 only on an \emph{Accepted} verdict, and Wrong Answer, Time/Memory Limit Exceeded (7.3\% and 0.9\% of analysed runs), Runtime Error, Compile Error, and responses without extractable code all score 0.

\paragraph{Models and decoding.} We evaluate 16 models from five providers under two prompt templates (\textsc{Detailed} and \textsc{Minimal}, Appendix~\ref{sec:prompts}) with $R{=}5$ runs per problem and prompt. Decoding was configured as temperature 0.3, top-$p$ 0.9, and a 4{,}096-token output limit; every request also carried the same one-sentence system message (Appendix~\ref{sec:prompts}). Provider APIs impose deviations, listed per model in Appendix~\ref{sec:model-snapshots}: Claude models received temperature only (the Anthropic API rejects both parameters) and returned code through a forced tool call; o4-mini accepts neither parameter; Gemini~2.5 Flash requests used a 16{,}384-token limit; 20 GPT-4.1 requests used a 2{,}048-token limit. All generations were obtained through the providers' batch APIs, one job per model and prompt, so the five runs of a problem are five independent requests in the same job: the run index is a request label, not a generation order, and provider-side processing order is unknown. Models were run on different days over 2026-03-23 to 2026-04-06. Appendix~\ref{sec:independence} checks for run-label effects.

\paragraph{Sample accounting.} The design calls for $16\times2\times100\times5=16{,}000$ requests. 15{,}995 returned a response and were judged; the five GPT-4.1 / \textsc{Detailed} requests for problem 3873 returned no response from the batch job (a response without extractable code would have been judged and scored 0). To keep a fully crossed design, that problem is dropped for every model and prompt, so all analyses use the $N{=}99$ problems common to every configuration: $16\times2\times99\times5=15{,}840$ judged runs. This complete-case exclusion removes one Hard problem and changes a model's run-level pass rate by at most 0.72\,pp. Two systems are excluded from pooled component statistics and flagged elsewhere: o4-mini (decoding parameters not user-controllable, possibly near-deterministic) and Gemini~2.5 Flash+Think (output budget differed across benchmark halves, as it also did for o4-mini); per-model numbers for all 16 appear in Appendix~\ref{sec:full-results}.

\section{Results}

\subsection{Leaderboards Collapse Under Correct Intervals}
\label{sec:collapse}
The natural-seeming analysis of a repeated-run experiment pools the $N\times R=495$ runs into one binomial proportion. Because runs within a problem share $p_i$, this interval answers neither estimand. Against the superpopulation target, a two-level bootstrap (resample problems, then runs within problem; Appendix~\ref{sec:bootstrap} gives the procedure and a simulation check of its coverage) is a median $2.05\times$ wider than the pooled Wilson interval (range 1.94--2.22 over the 14 clean systems; Table~\ref{tab:interval-widths}); the median ICC of 0.744 predicts a design effect of $\sqrt{1+4\rho}=1.99$, consistent with the measured 2.05. Against the fixed-benchmark target, the pooled interval errs the other way: the correct variance is $\mu_w/(NR)$, and the pooled interval is a median $1.96\times$ too \emph{wide}, because it mistakes persistent problem difficulty for run-level noise. One number, wrong in both directions depending on the question---which is why we report components rather than a single interval.

\begin{figure}[t]
\centering
\includegraphics[width=0.72\linewidth]{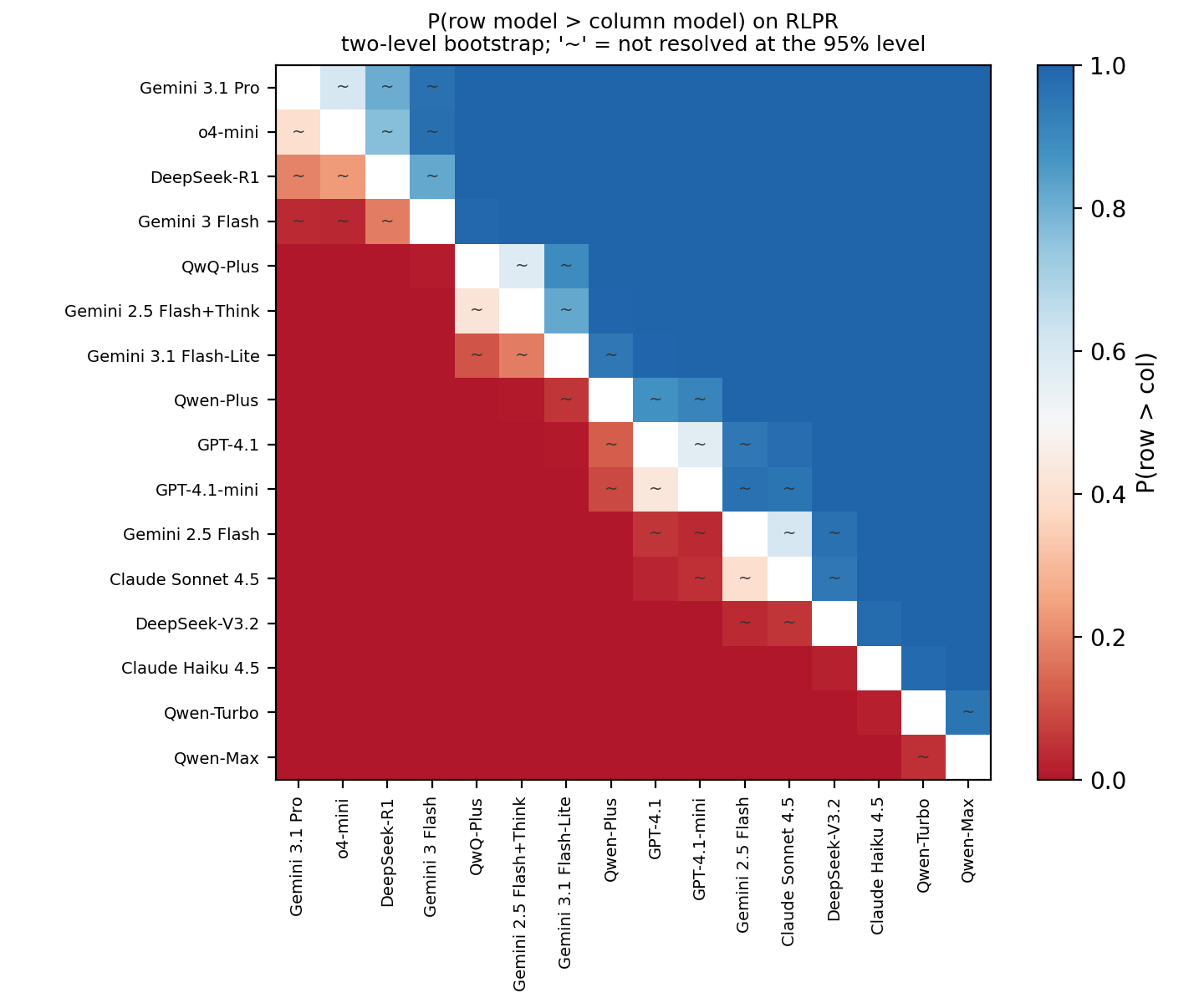}
\caption{Pairwise $P(\text{row}>\text{col})$ on run-level pass rate under the paired two-level bootstrap (problems resampled once for all models). ``$\sim$'' marks pairs unresolved at the 95\% level: 12 of 15 adjacent pairs. Rank intervals and top-$k$ probabilities for every model are in Appendix~\ref{sec:ranks}.}
\label{fig:rank}
\end{figure}

The practical consequence is shown in Figure~\ref{fig:rank}: under paired clustered bootstrap comparisons, 12 of the 15 adjacent pairs in the leaderboard---including the entire top three and the GPT-4.1 vs.\ GPT-4.1-mini comparison ($P=0.57$)---are unresolved at the 95\% level (13 of 15 in 5 of 50 bootstrap seeds); only three neighbours separate (Gemini 3 Flash vs.\ QwQ-Plus, DeepSeek-V3.2 vs.\ Claude Haiku 4.5, Claude Haiku 4.5 vs.\ Qwen-Turbo), and 95\% rank intervals span two to four positions for every model but one (Appendix~\ref{sec:ranks}). A leaderboard published from this protocol with pooled intervals would visually separate models that the data cannot distinguish.

\subsection{The Variance Ceiling, Measured}
\label{sec:ceiling}
\begin{figure}[t]
\centering
\includegraphics[width=0.66\linewidth]{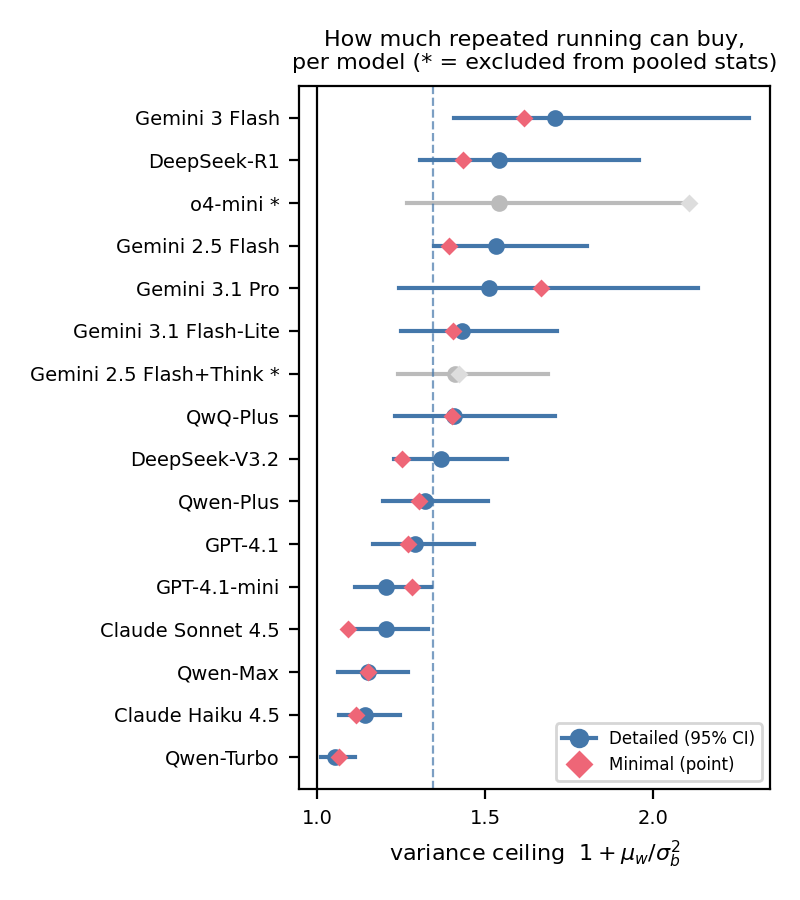}
\caption{Per-model variance ceiling $1+\mu_w/\sigma_b^2$ (Eq.~\ref{eq:ceiling}) with 95\% problem-bootstrap CIs (\textsc{Detailed}), and \textsc{Minimal}-prompt point estimates (diamonds). Dashed line: median of the 14 clean systems. Greyed models are excluded from pooled statistics (\S\ref{sec:setup}).}
\label{fig:forest}
\end{figure}

Figure~\ref{fig:forest} reports Eq.~\ref{eq:ceiling} per model with problem-bootstrap CIs; the underlying $\hat\mu_w$, $\widehat{\sigma_b^2}$, ICC, and ceiling estimates with intervals are tabulated for both prompts in Appendix~\ref{sec:full-results}. Across the 14 clean systems the ceiling has median 1.34, ranging from 1.05 (Qwen-Turbo) to 1.71 (Gemini 3 Flash); the aleatoric share of score variance at $R{=}5$ has median 6.4\% (range 1.0--12.4\%). Converted to interval widths, unlimited repetition can narrow a confidence interval by at most $1-1/\sqrt{1.34}\approx14\%$ for the median model and 23\% for the most decoding-sensitive one. The mechanism is visible in the raw outcomes: averaged over all 16 models, 79.7\% of problems are passed on all five runs or failed on all five (79.9\% over the 14 clean systems), so the intermediate regime that repetition averages over is thinly populated. The ceiling is also stable across our two prompt conditions (median per-model shift 0.05, max 0.16; diamonds in Figure~\ref{fig:forest})---and under random problem subsets of size 50 the 5--95\% range of the median ceiling is 1.25--1.44 (Appendix~\ref{sec:subsample}). Together with Appendix~\ref{sec:subsample} this is the only sensitivity evidence the study can offer, and no substitute for varying temperature (\S\ref{sec:scope}).

\subsection{Cost-Aware Allocation}
\label{sec:alloc}
Under a generation-only budget, Eq.~\ref{eq:var} times $N^{-1}\propto(R\sigma_b^2+\mu_w)$ is increasing in $R$, so $R{=}1$ maximises precision---but this assumes new verified problems are free, which for contamination-resistant coding benchmarks they are not. With the pooled ratio $\mu_w/\sigma_b^2=0.28$ (the ratio of the median components reported in \S\ref{sec:ceiling}), $R^{*}=0.53\sqrt{c_p/c_g}$: repetition starts paying once a problem costs more than $\approx3.5$ generations, and the $R{=}5$ design used here is optimal at $c_p/c_g\approx88$. Using the median of per-model ratios (0.34) instead gives $R^{*}=0.59\sqrt{c_p/c_g}$ and makes $R{=}5$ optimal at $c_p/c_g\approx73$; the conclusion does not depend on the choice.

More consequential than the optimum is the shape of the curve around it, which is flat on the side of over-repetition: if problems are ten times more expensive than assumed (so $R^{*}\approx16$), running $R{=}5$ costs 1.3\% in interval width; if they are ten times cheaper ($R^{*}\approx1.6$), $R{=}5$ costs 9\%. In both directions the ceiling (\S\ref{sec:ceiling}) caps what any $R$ can do. On this benchmark the load-bearing reason to run $R\geq2$ is therefore not precision but \emph{identification}---and not of the superpopulation interval, which a problem-level bootstrap of single runs already delivers (Appendix~\ref{sec:bootstrap}), but of $\sigma_b^2$, the ICC, the design effect, the fixed-benchmark interval, and the ceiling, none of which is estimable at $R{=}1$. In our data a pilot of $R'{=}2$--$3$ runs on half the problems estimates the ceiling on the other half with mean absolute error 0.18--0.27 and the ICC with 0.09--0.11, against floors of 0.16 and 0.08 set by problem sampling alone when all five runs are used; $\widehat{\sigma_b^2}$ was never negative (Appendix~\ref{sec:identification}).

\subsection{Scope: What These Measurements License}
\label{sec:scope}
These are measurements of one benchmark under one decoding configuration, and three limits follow. (i)~\emph{Temperature.} $\mu_w$ depends directly on decoding entropy; at $T{\to}0$ it approaches 0 and it is largest where $p_i\!\approx\!0.5$. Our $T{=}0.3$ sits below the $0.6$--$1.0$ range common in pass@$k$ work, so the ceilings in Figure~\ref{fig:forest} are specific to this configuration; prompt-stability is not temperature-stability. (ii)~\emph{Polarisation.} The small aleatoric share partly reflects this benchmark's difficulty polarisation; a benchmark with more mid-difficulty problems would have a higher ceiling. (iii)~\emph{Sampling.} Problems were taken by recency rank, not sampled from a defined population; the nearest definable population is ``public, non-premium LeetCode problems released in the window ending 2026-03-22'', and even for it the sample is a census of one window rather than a random draw, so the superpopulation reading of $\sigma_b^2$ is an idealisation---which is why \S\ref{sec:collapse} reports the fixed-benchmark reading alongside it. What travels is the method: the components, the ceiling, and $R^{*}$ are estimable from any repeated-run pilot at the operating temperature of interest.

\section{Recommendation and Conclusion}
For repeated-run evaluation of deterministic tasks we recommend: run a pilot with $R{=}2$--$3$; report $\widehat{\mu}$, $\widehat{\sigma_b^2}$, $\mathrm{AV}$, and the ICC alongside the score; compute intervals by problem-level (clustered) resampling, never by pooling runs; use Eq.~\ref{eq:ceiling} to decide whether further repetition is worth anything; and spend the remaining budget on problems. On our benchmark this accounting shows repetition can shrink intervals by at most 23\% (14\% for the median model), while the uncorrected pooled analysis misstates uncertainty by roughly $2\times$ in either direction and dissolves 12 of 15 adjacent leaderboard distinctions. In this regime, repetition's real product is not a sharper point estimate but the identification of the uncertainty structure itself; whether the same holds at the higher temperatures used in pass@$k$ evaluation is an open measurement, not a conclusion of this paper.

\section*{Limitations}
Beyond the scope limits of \S\ref{sec:scope}: the benchmark is small ($N{=}99$ analysed) and Python-only, and recency-based selection does not eliminate contamination; no contamination detection was performed. The variance model assumes conditionally independent Bernoulli runs given a problem, which provider-side caching or serving drift would violate; we did not instrument for this; because all runs were batch requests whose processing order is unknown, the post-hoc diagnostics in Appendix~\ref{sec:independence} can detect only effects tied to the run label, not drift over time or coupling that is constant across runs. Pooled quantities are medians over 14 systems after excluding two whose decoding regime or output budget was inconsistent (\S\ref{sec:setup}); per-model values vary widely (ceiling 1.05--1.71), so pooled numbers should not be applied to any single system. The moment estimator of $\sigma_b^2$ is truncated at zero and its beta-binomial refit uses $\alpha,\beta\!\ll\!1$ where the fit is sensitive (Appendix~\ref{sec:bb}); conclusions in the main text do not depend on the parametric fit. The verifier is LeetCode's hidden test suite, which is not exhaustive and not fully deterministic: a false accept adds a verifier-error component that the model in \S\ref{sec:sources} does not represent, and we have no augmented tests with which to bound it; and Time and Memory Limit Exceeded verdicts (7.3\% and 0.9\% of analysed runs) depend on judge timing and can differ between submissions of identical code, so some judge noise is absorbed into $\mu_w$ and inflates the aleatoric component. The cost model behind $R^{*}$ is linear with a single $c_p$ and $c_g$; real evaluation costs vary across problems, models, and infrastructure, so $R^{*}$ is illustrative rather than prescriptive. The two prompt templates differ only in instruction density, and with $N{=}99$ the paired design cannot detect prompt effects below $\approx$5\,pp at 80\% power; prompt claims here are limited to the stability of the ceiling, not the absence of prompt effects. Finally, all conclusions concern single-turn code generation with executable verification; multi-turn, tool-using, or non-deterministic-judge settings require separate treatment.

\section*{Ethics Statement}
This paper evaluates commercial and open-access LLM APIs on deterministic programming tasks. It involves no human subjects, personal data, or annotation labour. Evaluation relies on automated submission to an external online judge; such use should be understood as research benchmarking under the platform's operational constraints, and our data release accordingly contains binary outcomes, problem identifiers, and analysis code only, with no credentials, session artifacts, submission identifiers, or generated programs. Models were accessed through third-party APIs whose outputs and availability may change over time, so results are measurements of accessed systems under a documented protocol rather than immutable properties of a provider's model family. More broadly, stronger coding performance can increase both beneficial and harmful capabilities; we frame this work as evaluation methodology rather than an endorsement of autonomous code deployment.

\begin{ack}
This work received no external funding. All API usage and judge submissions were paid for by the authors; no author has a financial relationship with any of the model providers evaluated.
\end{ack}

\bibliographystyle{plainnat}
\bibliography{references}

\appendix

\section{Prompt Templates}
\label{sec:prompts}
Every request carried the system message ``You are an expert Python programmer specializing in algorithmic problem solving.'' (for Gemini~2.5 Flash, with the appended sentence ``Return only the Python code solution. Use \textbackslash n for newlines in the code string.'' and a JSON response schema with a single \texttt{code} field) followed by one of the two user messages below.

\subsection{Detailed Prompt}

\begin{lstlisting}
<role>
You are an expert Python algorithm engineer solving deterministic
LeetCode-style problems.
</role>

<task>
Produce a correct Python solution that fits the provided template and
is optimized for the problem constraints.
</task>

<instructions>
1. Read the full problem statement, constraints, and template before
   writing code.
2. Produce the best practical algorithm for the stated constraints,
   aiming for the optimal time complexity when possible.
3. Do not return a brute-force, quadratic, exponential, or placeholder
   solution when the constraints require a more efficient algorithm.
4. Use the required method signature from the template exactly.
5. Ensure the algorithm is complete and submission-ready, not a sketch
   or partial attempt.
6. Handle edge cases implied by the statement and constraints.
7. Return one complete Python solution that can be submitted directly.
</instructions>

<format_requirements>
Return plain Python source code only.
Start directly with the code.
Do not include explanations, markdown fences, or surrounding commentary.
The code must be valid, runnable Python 3 with no syntax errors.
The code must fit the provided code template without requiring manual edits.
The final code should be written for Accepted-style performance, not just
sample-case correctness.
Only include comments when they clarify non-obvious logic.
</format_requirements>

<problem>
{problem_description}
</problem>

<code_template>
{code_template}
</code_template>
\end{lstlisting}

\subsection{Minimal Prompt}

\begin{lstlisting}
<role>
You are a Python coding assistant solving a deterministic
LeetCode-style problem.
</role>

<task>
Write one correct and efficient Python solution that fits the provided
template.
</task>

<instructions>
1. Infer an algorithm that matches the constraints before writing code.
2. Prefer the optimal or near-optimal solution for the input limits.
3. Do not output brute-force code unless the constraints clearly make
   it acceptable.
4. Use the exact method signature from the template.
5. Return only final submission code.
</instructions>

<format_requirements>
Return plain Python source code only.
Start directly with the code.
Do not include explanations, markdown fences, or extra text.
The code must be valid, runnable Python 3 with no syntax errors.
The code must fit the provided code template without requiring manual edits.
The solution must be efficient enough to avoid obvious Time Limit
Exceeded outcomes under the given constraints.
</format_requirements>

<problem>
{problem_description}
</problem>

<code_template>
{code_template}
</code_template>
\end{lstlisting}

\section{Full Results and Fitted Dispersion}
\label{sec:full-results}

\begin{figure}[htbp]
\centering
\includegraphics[width=0.62\linewidth]{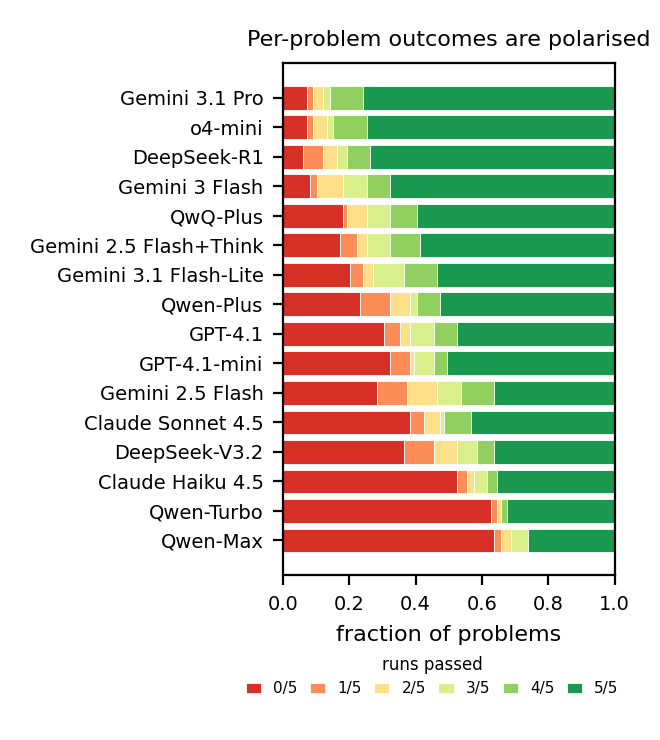}
\caption{Distribution of per-problem outcomes over $R=5$ runs (\textsc{Detailed}, models ordered by score). Averaged over models, 79.7\% of problems sit at $0/5$ or $5/5$ (range 64.6--94.9\%), which is why $\mu_w\ll\sigma_b^2$ on this benchmark.}
\label{fig:pol}
\end{figure}

Table~\ref{tab:full} reports both prompt conditions for all 16 models on the common $N=99$ problem set, together with the method-of-moments beta-binomial parameters $(\alpha,\beta)$ fitted to the \textsc{Detailed} condition. Tables~\ref{tab:components-detailed} and \ref{tab:components-minimal} give the variance components of \S\ref{sec:sources} per model and prompt with 95\% problem-bootstrap intervals; Table~\ref{tab:interval-widths} gives the three interval widths behind \S\ref{sec:collapse}. Under the \textsc{Minimal} prompt the 14-system medians are ICC 0.77 and ceiling 1.29 (range 1.06--1.67), against 0.74 and 1.34 under \textsc{Detailed}. Every table and figure in this paper is produced by a single script released with the data (footnote on page~1). Under the beta-binomial model, retry-free coverage is $\prod_{j=0}^{R-1}(\alpha+j)/(\alpha+\beta+j)$ and pass@$k$ is $1-\prod_{j=0}^{k-1}(\beta+j)/(\alpha+\beta+j)$, so the full retry curve follows from two numbers without further evaluation runs; Appendix~\ref{sec:bb} checks the internal consistency of this prediction. Note that $\alpha,\beta \ll 1$ throughout, the parametric signature of the difficulty polarisation shown in Figure~\ref{fig:pol}: the fitted Beta densities place their mass near $0$ and $1$ rather than near the mean.

\begin{table}[htbp]
\centering
\small
\caption{Full results on the common problem set ($N=99$, $R=5$), both prompt conditions. D $=$ \textsc{Detailed}, M $=$ \textsc{Minimal}. AV is $\times10^{-2}$ and uses the unbiased within-problem estimator $\frac{R}{R-1}\hat p_i(1-\hat p_i)$. $(\alpha,\beta)$ are moment-matched beta-binomial parameters for the \textsc{Detailed} condition. $\dagger$ reasoning-specialised. $\ddagger$ mixed output budget across benchmark halves (\S\ref{sec:setup}).}
\label{tab:full}
\begin{tabular}{lcccccccc}
\toprule
\multirow{2}{*}{Model} & \multicolumn{2}{c}{RLPR (\%)} & \multicolumn{2}{c}{PSR (\%)} & \multicolumn{2}{c}{AV ($\times10^{-2}$)} & \multicolumn{2}{c}{Beta fit (D)} \\
\cmidrule(lr){2-3}\cmidrule(lr){4-5}\cmidrule(lr){6-7}\cmidrule(lr){8-9}
 & D & M & D & M & D & M & $\alpha$ & $\beta$ \\
\midrule
GPT-4.1 & 59.6 & 59.8 & 47.5 & 49.5 & 5.45 & 5.15 & 0.17 & 0.11 \\
GPT-4.1-mini & 59.0 & 59.4 & 50.5 & 47.5 & 4.14 & 5.35 & 0.11 & 0.08 \\
o4-mini$^{\dagger\ddagger}$ & 86.1 & 85.5 & 74.7 & 68.7 & 4.24 & 6.57 & 0.46 & 0.07 \\
Claude Sonnet 4.5 & 53.3 & 51.9 & 43.4 & 47.5 & 4.24 & 2.12 & 0.10 & 0.09 \\
Claude Haiku 4.5 & 41.6 & 41.2 & 35.4 & 34.3 & 3.03 & 2.53 & 0.05 & 0.08 \\
Gemini 3.1 Pro & 86.7 & 85.7 & 75.8 & 73.7 & 3.94 & 4.95 & 0.43 & 0.07 \\
Gemini 3 Flash & 81.2 & 83.2 & 67.7 & 70.7 & 6.36 & 5.35 & 0.56 & 0.13 \\
Gemini 3.1 Flash-Lite & 69.1 & 64.0 & 53.5 & 50.5 & 6.46 & 6.67 & 0.29 & 0.13 \\
Gemini 2.5 Flash & 54.1 & 48.3 & 36.4 & 33.3 & 8.69 & 7.07 & 0.28 & 0.24 \\
Gemini 2.5 Flash+Think$^{\dagger\ddagger}$ & 72.3 & 74.5 & 58.6 & 57.6 & 5.86 & 5.66 & 0.29 & 0.11 \\
QwQ-Plus$^{\dagger}$ & 72.9 & 72.5 & 59.6 & 58.6 & 5.76 & 5.76 & 0.29 & 0.11 \\
Qwen-Plus & 63.6 & 61.4 & 52.5 & 51.5 & 5.66 & 5.56 & 0.20 & 0.11 \\
Qwen-Max & 30.9 & 30.7 & 26.3 & 25.3 & 2.83 & 2.83 & 0.04 & 0.10 \\
Qwen-Turbo & 34.7 & 31.1 & 32.3 & 28.3 & 1.11 & 1.31 & 0.01 & 0.03 \\
DeepSeek-R1$^{\dagger}$ & 84.0 & 84.2 & 73.7 & 74.7 & 4.75 & 4.04 & 0.45 & 0.08 \\
DeepSeek-V3.2 & 48.7 & 48.1 & 36.4 & 38.4 & 6.77 & 5.05 & 0.17 & 0.18 \\
\bottomrule
\end{tabular}
\end{table}

\input{tables/tab_components_detailed}
\input{tables/tab_components_minimal}
\input{tables/tab_interval_widths}

\section{Predicting Retry Curves from Dispersion}
\label{sec:bb}

\begin{figure}[htbp]
\centering
\includegraphics[width=0.6\linewidth]{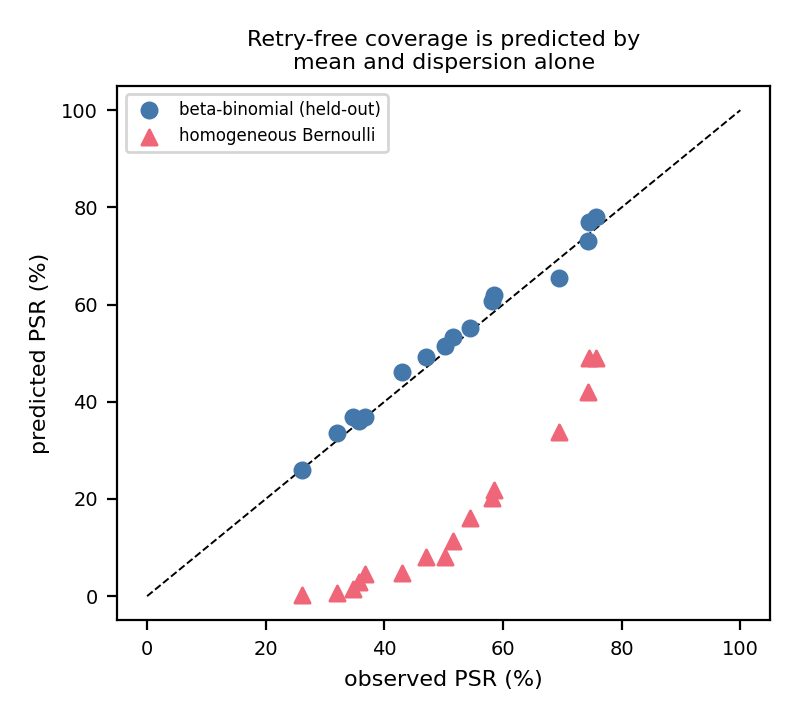}
\caption{Held-out prediction of retry-free coverage from the two-parameter beta-binomial (circles; each point is one model, prediction and observation averaged over 50 half-splits). The homogeneous-Bernoulli points (triangles) are shown only as a structural floor, not as a serious baseline.}
\label{fig:bb-calib}
\end{figure}

The components $(\mu,\sigma_b^2)$ do more than support inference on the mean: matched to a $\mathrm{Beta}(\alpha,\beta)$, they predict deployment-relevant functionals of the $p_i$ distribution. To test this without fitting and evaluating on the same problems, we use 50 random half-splits per model: estimate $(\alpha,\beta)$ on half the problems, predict on the held-out half. Two error measures must be kept apart. \emph{Per split}, the held-out retry-free coverage (fraction of problems passed on all five runs) is predicted with mean absolute error 7.9\,pp and held-out pass@3 with 6.4\,pp---but simply carrying over the estimation half's observed coverage misses the held-out half by 7.7\,pp, so per split the parametric model adds no skill beyond the observed value: the error is dominated by problem-sampling noise between two halves of 50 and 49 problems. \emph{Averaged over the 50 splits per model}, which removes that noise and is close to an in-sample check (each problem sits in the estimation half about half the time), the predictions are nearly unbiased: mean absolute bias 1.6\,pp for retry-free coverage and 1.0\,pp for pass@3 (ranges over 20 bootstrap seeds 1.1--2.0 and 0.7--1.3; Figure~\ref{fig:bb-calib} plots the split-averaged predictions at the released seed, 1.8 and 0.7\,pp).\footnote{An earlier version of this paper reported 1.46 and 1.07\,pp without stating that these were split-averaged values.} What the two-parameter fit adds is therefore not per-split accuracy on a new problem set but the whole pass@$k$ curve from two numbers, with no systematic bias.

A fair baseline must be one a practitioner might actually use. Predicting a model's retry-free coverage by leave-one-model-out linear regression on its run-level pass rate (using the other 15 models to fit the line) achieves 2.89\,pp MAE---closer than the homogeneous-Bernoulli floor (predicting $\mathrm{RLPR}^5$; $\approx$34\,pp, the triangles in Figure~\ref{fig:bb-calib}, shown only for reference), but above the split-averaged beta-binomial bias, and unlike the beta-binomial it predicts a single scalar rather than the full pass@$k$ curve and requires a fleet of comparison models. Two caveats: with $\alpha,\beta\ll1$ the moment-matched fit is sensitive to $\widehat{\sigma_b^2}$, and the half-split validation spans the 99 collected problems rather than a broader task population, so this is internal consistency, not external generalisation.

Table~\ref{tab:repeats} shows how estimation degrades when dispersion is estimated from fewer runs \emph{on the same problems}: subsets of $R'{=}2$ runs predict the five-run retry-free coverage to 2.30\,pp and $R'{=}3$ to 1.50\,pp, against the 1.00\,pp in-sample floor of the parametric family itself. Because the $R'$ runs are among the five that define the target, this is an in-sample measure of how much dispersion information $R'$ runs carry; the out-of-sample version, with disjoint problems and runs, is Appendix~\ref{sec:identification}.

\begin{table}[htbp]
\centering
\small
\caption{Mean absolute error when dispersion is estimated from only $R'$ of the 5 runs (all $\binom{5}{R'}$ subsets, 16 models) and used to predict five-run retry-free coverage on the same problems. The $R'{=}5$ row is the in-sample approximation error of the beta-binomial family, not zero.}
\label{tab:repeats}
\begin{tabular}{lcccc}
\toprule
$R'$ & 2 & 3 & 4 & 5 \\
\midrule
PSR error (pp) & 2.30 & 1.50 & 1.13 & 1.00 \\
$\widehat{\sigma_b^2}$ error & .0096 & .0054 & .0032 & --- \\
\bottomrule
\end{tabular}
\end{table}

\FloatBarrier
\section{Interval Procedures and Their Coverage}
\label{sec:bootstrap}
\input{tables/alg_bootstrap}
\input{tables/tab_coverage}

\FloatBarrier
\section{Identification from Two or Three Runs, Out of Sample}
\label{sec:identification}
Table~\ref{tab:repeats} in Appendix~\ref{sec:bb} estimates dispersion from $R'$ runs and evaluates it on the same problems and runs, so predictor and target are dependent. Table~\ref{tab:repeats-disjoint} removes the dependence: for each of 200 random splits per model, the components are estimated from $R'$ randomly chosen runs on a random half of the problems and compared with reference values computed on the other half from the runs not used for estimation. The $R'{=}5$ row, which uses all five runs on both halves, is the floor set by problem-sampling noise between two halves of 50 and 49 problems. At $R'{=}2$ the ceiling is recovered with mean absolute error 0.25--0.27 (the two reference definitions) and the ICC with 0.10--0.11; at $R'{=}3$ with 0.18--0.25 and 0.09--0.11; the floors are 0.16 and 0.08. Relative to the median ceiling excess $\mu_w/\sigma_b^2\approx0.34$ these are large errors, which is why the recommendation is a pilot, not a measurement. The moment estimator $\widehat{\sigma_b^2}$ of the estimation half was positive in every one of the 2{,}800 splits at every $R'$ (the reference half gave $\widehat{\sigma_b^2}\le0$ in 3 splits at $R'{=}3$), and in 99.99\% of bootstrap replicates at $R'{=}2$, so on this benchmark the components are identifiable from two runs; what two runs do not give is a tight interval on the ceiling: the median 95\% CI width is 0.54 at $R'{=}2$, 0.39 at $R'{=}3$, and 0.33 at $R'{=}5$. This is the evidence behind the recommendation in \S\ref{sec:alloc}: two to three runs identify the uncertainty structure; going from three runs to five narrows the interval on the ceiling by a further 15\%.
\input{tables/tab_repeats_disjoint}

\FloatBarrier
\section{Rank Uncertainty}
\label{sec:ranks}
Figure~\ref{fig:rank} reports only pairwise probabilities. Table~\ref{tab:ranks} gives, under the same paired two-level bootstrap, each model's 95\% rank interval and its probability of finishing first or in the top three. The top-1 probability is split between Gemini 3.1 Pro (0.57), o4-mini (0.35), and DeepSeek-R1 (0.11) (the three sum to more than one because tied replicates share the better rank); the middle of the table is where rank intervals are widest, and only Claude Haiku 4.5 has a rank that the data pin down exactly. We do not apply a multiplicity correction to the 15 adjacent comparisons, since the purpose is to show how little the data separate, and a correction would only enlarge the unresolved set.
\input{tables/tab_ranks}

\FloatBarrier
\section{Run-Label Diagnostics}
\label{sec:independence}
The variance model of \S\ref{sec:sources} assumes that, given $p_i$, the five runs on a problem are independent. Because every run was a separate request in one provider batch job (\S\ref{sec:setup}), the run index is a label with no known temporal meaning, and the two checks possible with these data (Table~\ref{tab:independence}) are correspondingly limited. First, label effects: if the provider had treated the request slots differently, pass rates would differ systematically by run index. They do not: the largest gap between run indices is 0.8\,pp (\textsc{Detailed}) and 1.4\,pp (\textsc{Minimal}), and Cochran's $Q$ summed over the 14 clean systems gives $p=0.90$ and $p=0.56$, with no individual model below $p=0.079$. Second, excess agreement between index-adjacent runs: among problems with mixed outcomes, runs with adjacent indices agree no more often than runs with non-adjacent indices (49.7\% vs.\ 51.9\%, permutation $p=0.09$; the small difference is in the direction opposite to positive dependence). Neither check can detect drift over the hours a batch job takes, nor coupling that is constant across all five runs of a problem, such as a cached response served every time; such coupling would appear as $\mu_w\approx0$ and an ICC near 1, which only Qwen-Turbo approaches. A protocol that wants to test conditional independence should timestamp requests or interleave them deliberately; ours did not.
\input{tables/tab_independence}

\FloatBarrier
\section{Ceiling Stability Under Problem Subsampling}
\label{sec:subsample}
To quantify how much the measured ceiling depends on which problems happened to be collected, Table~\ref{tab:ceiling-subsample} recomputes the per-model ceiling on random subsets of 30, 50, and 70 of the 99 problems. The 5--95\% range of the 14-system median ceiling is 1.20--1.45 at $n=30$ and 1.28--1.40 at $n=70$; the models at the extremes keep their order (Qwen-Turbo's 5--95\% range at $n=50$ is 1.02--1.09, Gemini 3 Flash's 1.44--2.24), and at $n=30$ 0.6\% of draws gave $\widehat{\sigma_b^2}\le0$ and hence an undefined ceiling. Because subsets are drawn without replacement from the same 99 problems, large subsets overlap heavily and this understates the variation between independent benchmarks of that size. The level of the median ceiling is a property of this benchmark's difficulty mix as a whole rather than of a few problems, but individual models' ceilings are far less stable than the median.
\input{tables/tab_ceiling_subsample}

\FloatBarrier
\section{Model Snapshots and Access Paths}
\label{sec:model-snapshots}
Table~\ref{tab:model-snapshots} lists the model identifiers and access dates. All requests went through batch endpoints: OpenAI Batch API (GPT-4.1, GPT-4.1-mini, o4-mini), Anthropic Message Batches (Claude), Google Vertex AI batch prediction (Gemini), and Alibaba DashScope batch (Qwen and DeepSeek; the DeepSeek models were DashScope-hosted). Parameters actually sent, per client: OpenAI GPT-4.1 models---temperature 0.3, top-$p$ 0.9, \texttt{max\_tokens} 4{,}096 (2{,}048 for the 20 requests on problems 3759 and 3785); o4-mini---\texttt{max\_completion\_tokens} only (4{,}096 for one benchmark half, 16{,}384 for the other), no temperature or top-$p$. Anthropic---temperature 0.3, \texttt{max\_tokens} 4{,}096, no top-$p$, a forced \texttt{submit\_solution} tool call returning the code string. Gemini 3.x (run 2026-03-28 to 03-30)---temperature 0.3, top-$p$ 0.9, \texttt{maxOutputTokens} 4{,}096, provider-default thinking, plain-text output; Gemini 2.5 Flash (2026-04-05)---temperature 0.3, top-$p$ 0.9, \texttt{maxOutputTokens} 16{,}384, JSON-schema output, and a \texttt{thinkingLevel: LOW} setting that is defined for Gemini~3 and whose effect on a 2.5 model we cannot verify from our logs; Gemini 2.5 Flash+Think---the same with \texttt{thinkingBudget} 8{,}192 and \texttt{maxOutputTokens} 4{,}096 for one benchmark half and 24{,}576 for the other. DashScope---temperature 0.3, top-$p$ 0.9, \texttt{max\_tokens} 4{,}096. The per-generation logs record configured rather than sent values (e.g.\ they list top-$p$ 0.9 for Claude), which is why this paragraph is derived from the client code at the commits in use on each run date.

\begin{table}[htbp]
\centering
\small
\caption{Model snapshots used in the experiments. The API identifier column records the concrete model string used by the evaluation pipeline. Access dates are derived from the archived run dates for the corresponding completed paper runs. $^{\S}$ accessed through Alibaba DashScope.}
\label{tab:model-snapshots}
\begin{tabular}{llll}
\toprule
Display Name & Provider & API Identifier & Access Date \\
\midrule
GPT-4.1 & OpenAI & gpt-4.1 & 2026-04-01 \\
GPT-4.1-mini & OpenAI & gpt-4.1-mini & 2026-03-31 \\
o4-mini & OpenAI & o4-mini & 2026-04-06 \\
Claude Sonnet 4.5 & Anthropic & claude-sonnet-4-5-20250929 & 2026-04-03 \\
Claude Haiku 4.5 & Anthropic & claude-haiku-4-5-20251001 & 2026-04-03 \\
Gemini 3.1 Pro (preview) & Google & gemini-3.1-pro-preview & 2026-03-28 \\
Gemini 3 Flash (preview) & Google & gemini-3-flash-preview & 2026-03-29 \\
Gemini 3.1 Flash-Lite (preview) & Google & gemini-3.1-flash-lite-preview & 2026-03-30 \\
Gemini 2.5 Flash & Google & gemini-2.5-flash & 2026-04-05 \\
Gemini 2.5 Flash+Think & Google & gemini-2.5-flash (thinkingBudget=8192) & 2026-04-05 \\
QwQ-Plus & Qwen & qwq-plus & 2026-03-23 \\
Qwen-Plus & Qwen & qwen-plus & 2026-03-29 \\
Qwen-Max & Qwen & qwen-max & 2026-03-23 \\
Qwen-Turbo & Qwen & qwen-turbo & 2026-03-23 \\
DeepSeek-R1 & DeepSeek$^{\S}$ & deepseek-r1 & 2026-03-24 \\
DeepSeek-V3.2 & DeepSeek$^{\S}$ & deepseek-v3.2 & 2026-03-25 \\
\bottomrule
\end{tabular}
\end{table}

\end{document}

%% file: tables/tab_components_detailed.tex
\begin{table}[htbp]
\centering
\small
\setlength{\tabcolsep}{4pt}
\caption{Per-model variance components, \textsc{Detailed} prompt ($N=99$, $R=5$), with 95\% percentile problem-bootstrap intervals ($B=2000$; problems resampled with their five runs). $\hat\mu_w=\mathrm{AV}$ and $\widehat{\sigma_b^2}=s^2(\hat p)-\mathrm{AV}/R$ are $\times10^{-2}$; ICC $=\sigma_b^2/(\sigma_b^2+\mu_w)$; ceiling $=1+\mu_w/\sigma_b^2$. No point estimate of $\sigma_b^2$ required truncation. Median over the 14 clean systems: ICC 0.744, ceiling 1.34. $\dagger$ reasoning-specialised; $\ddagger$ excluded from pooled statistics.}
\label{tab:components-detailed}
\begin{tabular}{lcccc}
\toprule
Model & $\hat\mu_w$ ($\times10^{-2}$) & $\widehat{\sigma_b^2}$ ($\times10^{-2}$) & ICC & Ceiling \\
\midrule
GPT-4.1 & 5.45 {\scriptsize[3.33, 7.68]} & 18.83 {\scriptsize[15.56, 21.10]} & 0.78 {\scriptsize[0.68, 0.86]} & 1.29 {\scriptsize[1.17, 1.47]} \\
GPT-4.1-mini & 4.14 {\scriptsize[2.42, 6.06]} & 20.26 {\scriptsize[17.42, 22.30]} & 0.83 {\scriptsize[0.75, 0.90]} & 1.20 {\scriptsize[1.11, 1.34]} \\
o4-mini$^{\dagger\ddagger}$ & 4.24 {\scriptsize[2.42, 6.06]} & 7.84 {\scriptsize[4.07, 11.72]} & 0.65 {\scriptsize[0.48, 0.79]} & 1.54 {\scriptsize[1.27, 2.10]} \\
Claude Sonnet 4.5 & 4.24 {\scriptsize[2.53, 6.16]} & 20.87 {\scriptsize[18.45, 22.55]} & 0.83 {\scriptsize[0.75, 0.90]} & 1.20 {\scriptsize[1.11, 1.33]} \\
Claude Haiku 4.5 & 3.03 {\scriptsize[1.41, 4.75]} & 21.49 {\scriptsize[18.79, 23.23]} & 0.88 {\scriptsize[0.80, 0.94]} & 1.14 {\scriptsize[1.06, 1.24]} \\
Gemini 3.1 Pro & 3.94 {\scriptsize[2.22, 5.66]} & 7.70 {\scriptsize[3.84, 11.52]} & 0.66 {\scriptsize[0.47, 0.81]} & 1.51 {\scriptsize[1.24, 2.13]} \\
Gemini 3 Flash & 6.36 {\scriptsize[4.24, 8.59]} & 9.00 {\scriptsize[5.36, 12.35]} & 0.59 {\scriptsize[0.44, 0.71]} & 1.71 {\scriptsize[1.41, 2.28]} \\
Gemini 3.1 Flash-Lite & 6.46 {\scriptsize[4.34, 8.69]} & 15.06 {\scriptsize[11.06, 18.19]} & 0.70 {\scriptsize[0.58, 0.80]} & 1.43 {\scriptsize[1.25, 1.71]} \\
Gemini 2.5 Flash & 8.69 {\scriptsize[6.36, 11.01]} & 16.32 {\scriptsize[13.61, 18.52]} & 0.65 {\scriptsize[0.55, 0.74]} & 1.53 {\scriptsize[1.35, 1.80]} \\
Gemini 2.5 Flash+Think$^{\dagger\ddagger}$ & 5.86 {\scriptsize[3.84, 7.98]} & 14.31 {\scriptsize[10.47, 17.69]} & 0.71 {\scriptsize[0.59, 0.81]} & 1.41 {\scriptsize[1.24, 1.69]} \\
QwQ-Plus$^{\dagger}$ & 5.76 {\scriptsize[3.74, 7.98]} & 14.14 {\scriptsize[9.91, 17.47]} & 0.71 {\scriptsize[0.59, 0.81]} & 1.41 {\scriptsize[1.23, 1.71]} \\
Qwen-Plus & 5.66 {\scriptsize[3.73, 7.78]} & 17.67 {\scriptsize[14.35, 20.11]} & 0.76 {\scriptsize[0.66, 0.84]} & 1.32 {\scriptsize[1.19, 1.51]} \\
Qwen-Max & 2.83 {\scriptsize[1.21, 4.55]} & 18.72 {\scriptsize[14.77, 21.66]} & 0.87 {\scriptsize[0.79, 0.94]} & 1.15 {\scriptsize[1.06, 1.27]} \\
Qwen-Turbo & 1.11 {\scriptsize[0.20, 2.32]} & 21.78 {\scriptsize[18.15, 24.03]} & 0.95 {\scriptsize[0.90, 0.99]} & 1.05 {\scriptsize[1.01, 1.11]} \\
DeepSeek-R1$^{\dagger}$ & 4.75 {\scriptsize[2.93, 6.77]} & 8.76 {\scriptsize[5.13, 12.33]} & 0.65 {\scriptsize[0.51, 0.77]} & 1.54 {\scriptsize[1.31, 1.96]} \\
DeepSeek-V3.2 & 6.77 {\scriptsize[4.65, 8.99]} & 18.41 {\scriptsize[15.88, 20.42]} & 0.73 {\scriptsize[0.64, 0.81]} & 1.37 {\scriptsize[1.23, 1.57]} \\
\bottomrule
\end{tabular}
\end{table}

%% file: tables/tab_components_minimal.tex
\begin{table}[htbp]
\centering
\small
\setlength{\tabcolsep}{4pt}
\caption{Per-model variance components, \textsc{Minimal} prompt ($N=99$, $R=5$), with 95\% percentile problem-bootstrap intervals ($B=2000$; problems resampled with their five runs). $\hat\mu_w=\mathrm{AV}$ and $\widehat{\sigma_b^2}=s^2(\hat p)-\mathrm{AV}/R$ are $\times10^{-2}$; ICC $=\sigma_b^2/(\sigma_b^2+\mu_w)$; ceiling $=1+\mu_w/\sigma_b^2$. No point estimate of $\sigma_b^2$ required truncation. Median over the 14 clean systems: ICC 0.774, ceiling 1.29. $\dagger$ reasoning-specialised; $\ddagger$ excluded from pooled statistics.}
\label{tab:components-minimal}
\begin{tabular}{lcccc}
\toprule
Model & $\hat\mu_w$ ($\times10^{-2}$) & $\widehat{\sigma_b^2}$ ($\times10^{-2}$) & ICC & Ceiling \\
\midrule
GPT-4.1 & 5.15 {\scriptsize[3.33, 7.27]} & 19.09 {\scriptsize[15.75, 21.12]} & 0.79 {\scriptsize[0.69, 0.86]} & 1.27 {\scriptsize[1.16, 1.45]} \\
GPT-4.1-mini & 5.35 {\scriptsize[3.43, 7.58]} & 18.97 {\scriptsize[15.82, 21.14]} & 0.78 {\scriptsize[0.69, 0.86]} & 1.28 {\scriptsize[1.16, 1.46]} \\
o4-mini$^{\dagger\ddagger}$ & 6.57 {\scriptsize[4.34, 8.79]} & 5.94 {\scriptsize[2.98, 9.03]} & 0.47 {\scriptsize[0.31, 0.62]} & 2.11 {\scriptsize[1.62, 3.20]} \\
Claude Sonnet 4.5 & 2.12 {\scriptsize[0.91, 3.64]} & 23.08 {\scriptsize[21.08, 24.24]} & 0.92 {\scriptsize[0.86, 0.96]} & 1.09 {\scriptsize[1.04, 1.17]} \\
Claude Haiku 4.5 & 2.53 {\scriptsize[1.21, 4.14]} & 21.93 {\scriptsize[19.31, 23.52]} & 0.90 {\scriptsize[0.83, 0.95]} & 1.12 {\scriptsize[1.05, 1.20]} \\
Gemini 3.1 Pro & 4.95 {\scriptsize[3.03, 6.97]} & 7.42 {\scriptsize[3.97, 10.96]} & 0.60 {\scriptsize[0.43, 0.74]} & 1.67 {\scriptsize[1.35, 2.31]} \\
Gemini 3 Flash & 5.35 {\scriptsize[3.23, 7.58]} & 8.70 {\scriptsize[4.85, 12.39]} & 0.62 {\scriptsize[0.45, 0.76]} & 1.62 {\scriptsize[1.32, 2.22]} \\
Gemini 3.1 Flash-Lite & 6.67 {\scriptsize[4.55, 8.89]} & 16.54 {\scriptsize[13.23, 19.06]} & 0.71 {\scriptsize[0.61, 0.80]} & 1.40 {\scriptsize[1.25, 1.63]} \\
Gemini 2.5 Flash & 7.07 {\scriptsize[4.95, 9.19]} & 18.10 {\scriptsize[15.65, 20.12]} & 0.72 {\scriptsize[0.63, 0.80]} & 1.39 {\scriptsize[1.24, 1.59]} \\
Gemini 2.5 Flash+Think$^{\dagger\ddagger}$ & 5.66 {\scriptsize[3.73, 7.58]} & 13.47 {\scriptsize[9.03, 17.15]} & 0.70 {\scriptsize[0.58, 0.81]} & 1.42 {\scriptsize[1.24, 1.73]} \\
QwQ-Plus$^{\dagger}$ & 5.76 {\scriptsize[3.74, 7.78]} & 14.32 {\scriptsize[10.19, 17.59]} & 0.71 {\scriptsize[0.60, 0.81]} & 1.40 {\scriptsize[1.23, 1.66]} \\
Qwen-Plus & 5.56 {\scriptsize[3.54, 7.78]} & 18.34 {\scriptsize[15.21, 20.63]} & 0.77 {\scriptsize[0.67, 0.85]} & 1.30 {\scriptsize[1.18, 1.48]} \\
Qwen-Max & 2.83 {\scriptsize[1.41, 4.65]} & 18.64 {\scriptsize[14.68, 21.58]} & 0.87 {\scriptsize[0.78, 0.93]} & 1.15 {\scriptsize[1.07, 1.27]} \\
Qwen-Turbo & 1.31 {\scriptsize[0.30, 2.53]} & 20.33 {\scriptsize[16.28, 23.19]} & 0.94 {\scriptsize[0.88, 0.99]} & 1.06 {\scriptsize[1.01, 1.13]} \\
DeepSeek-R1$^{\dagger}$ & 4.04 {\scriptsize[2.22, 6.06]} & 9.34 {\scriptsize[5.40, 13.33]} & 0.70 {\scriptsize[0.53, 0.82]} & 1.43 {\scriptsize[1.22, 1.88]} \\
DeepSeek-V3.2 & 5.05 {\scriptsize[3.13, 7.17]} & 20.13 {\scriptsize[17.61, 21.89]} & 0.80 {\scriptsize[0.71, 0.87]} & 1.25 {\scriptsize[1.15, 1.40]} \\
\bottomrule
\end{tabular}
\end{table}

%% file: tables/tab_interval_widths.tex
\begin{table}[htbp]
\centering
\small
\caption{Width (pp) of four 95\% intervals for the \textsc{Detailed} score: (a) pooled Wilson on $NR=495$ runs; (b) two-level bootstrap (problems, then runs within problem); (b$'$) one-level problem bootstrap; (c) fixed-benchmark normal interval with $\mathrm{Var}=\hat\mu_w/(NR)$. $B=2000$.}
\label{tab:interval-widths}
\begin{tabular}{lcccccc}
\toprule
Model & (a) & (b) & (b$'$) & (c) & (b)/(a) & (a)/(c) \\
\midrule
GPT-4.1 & 8.6 & 18.2 & 17.8 & 4.1 & 2.11 & 2.09 \\
GPT-4.1-mini & 8.6 & 18.2 & 17.8 & 3.6 & 2.11 & 2.41 \\
o4-mini$^{\dagger\ddagger}$ & 6.1 & 12.5 & 11.5 & 3.6 & 2.05 & 1.68 \\
Claude Sonnet 4.5 & 8.8 & 18.6 & 18.8 & 3.6 & 2.12 & 2.41 \\
Claude Haiku 4.5 & 8.7 & 18.6 & 18.8 & 3.1 & 2.15 & 2.82 \\
Gemini 3.1 Pro & 6.0 & 11.9 & 11.3 & 3.5 & 1.99 & 1.71 \\
Gemini 3 Flash & 6.9 & 13.3 & 12.5 & 4.4 & 1.94 & 1.55 \\
Gemini 3.1 Flash-Lite & 8.1 & 16.2 & 15.6 & 4.5 & 1.99 & 1.81 \\
Gemini 2.5 Flash & 8.7 & 17.0 & 16.8 & 5.2 & 1.94 & 1.68 \\
Gemini 2.5 Flash+Think$^{\dagger\ddagger}$ & 7.9 & 15.8 & 15.0 & 4.3 & 2.00 & 1.84 \\
QwQ-Plus$^{\dagger}$ & 7.8 & 15.4 & 15.6 & 4.2 & 1.97 & 1.85 \\
Qwen-Plus & 8.4 & 17.4 & 16.6 & 4.2 & 2.06 & 2.02 \\
Qwen-Max & 8.1 & 17.8 & 17.6 & 3.0 & 2.19 & 2.74 \\
Qwen-Turbo & 8.4 & 18.6 & 19.0 & 1.9 & 2.22 & 4.50 \\
DeepSeek-R1$^{\dagger}$ & 6.4 & 12.7 & 11.9 & 3.8 & 1.97 & 1.68 \\
DeepSeek-V3.2 & 8.8 & 18.0 & 17.2 & 4.6 & 2.05 & 1.91 \\
\midrule
Median (14 clean) & & & & & 2.05 & 1.96 \\
Range (14 clean) & & & & & 1.94--2.22 & 1.55--4.50 \\
\bottomrule
\end{tabular}
\end{table}

%% file: tables/alg_bootstrap.tex
\paragraph{Interval procedures.} Input: binary outcomes $Y_{i,r}$, $i=1,\dots,N$, $r=1,\dots,R$; $\hat p_i=\frac1R\sum_r Y_{i,r}$; $\widehat{\mu}=\frac1N\sum_i\hat p_i$.
\begin{enumerate}\itemsep1pt
\item[(a)] \emph{Pooled Wilson (shown for reference only).} $k=\sum_{i,r}Y_{i,r}$, $n=NR$; Wilson score interval for $k/n$. It treats the $NR$ runs as independent and targets neither estimand.
\item[(b)] \emph{Superpopulation interval (two-level clustered bootstrap).} For $b=1,\dots,B$ ($B=2000$):
  \begin{enumerate}\itemsep0pt
  \item[1.] draw problem indices $i^{*}_1,\dots,i^{*}_N$ uniformly with replacement from $\{1,\dots,N\}$;
  \item[2.] for each $j$, draw $R$ runs with replacement from the $R$ runs of problem $i^{*}_j$, i.e.\ $K_j\sim\mathrm{Binomial}(R,\hat p_{i^{*}_j})$;
  \item[3.] $S^{*}_b=\frac{1}{NR}\sum_{j=1}^{N}K_j$.
  \end{enumerate}
  Report the 2.5th and 97.5th percentiles of $\{S^{*}_b\}$. Target: $\mu=\mathbb{E}[p_i]$ over the task population. For paired model comparisons the same $i^{*}$ draw is shared by all models in a replicate. One-level variant (b$'$): omit step 2 and set $S^{*}_b=\frac1N\sum_j\hat p_{i^{*}_j}$.
\item[(c)] \emph{Fixed-benchmark interval.} Target: $\bar p=\frac1N\sum_i p_i$ for the $N$ problems at hand. $\widehat{\mathrm{Var}}=\mathrm{AV}/(NR)$ with $\mathrm{AV}=\frac1N\sum_i\frac{R}{R-1}\hat p_i(1-\hat p_i)$; interval $\widehat{\mu}\pm1.96\sqrt{\widehat{\mathrm{Var}}}$. (Bootstrap variant (c$'$): keep the problems fixed, draw $K_i\sim\mathrm{Binomial}(R,\hat p_i)$ for every $i$, and take the percentile interval of $\frac{1}{NR}\sum_i K_i$; it uses the biased $\hat p_i(1-\hat p_i)$ and is narrower by about $\sqrt{(R-1)/R}$.)
\end{enumerate}
\paragraph{Does (b) double-count run noise?} Resampling $\hat p_{i^{*}}$ already carries the finite-$R$ noise, since $\mathbb{E}\,s^2(\hat p)=\sigma_b^2+\mu_w/R$, so step 2 adds about $\frac{R-1}{R}\cdot\frac{\mu_w}{R}$ of within-problem variance a second time. Because $\mu_w/R\ll\sigma_b^2$ on this benchmark, the inflation is small: 0.3--4.8\% in width for the simulated configurations. It roughly offsets the slight narrowness of the percentile bootstrap at $N=99$. In simulation (Table~\ref{tab:coverage}), (b) covers $\mu$ in 94.9--95.6\% of datasets and (b$'$) in 93.5--94.8\%, against a nominal 95\%; (c) covers $\bar p$ in 91.3--94.5\%. We use (b) for all superpopulation intervals and rank comparisons.

%% file: tables/tab_coverage.tex
\begin{table}[htbp]
\centering
\small
\setlength{\tabcolsep}{2pt}
\caption{Empirical coverage (\%) of nominal 95\% intervals over 1000 simulated datasets ($N=99$, $R=5$, $p_i\sim\mathrm{Beta}$ moment-matched to the named model's \textsc{Detailed} components; the ICC quoted is that of the matched Beta). $\mu$: superpopulation mean; $\bar p=\frac1N\sum_i p_i$: realised fixed-benchmark mean; width in pp. Monte Carlo s.e.\ at 95\% is 0.7\,pp.}
\label{tab:coverage}
\begin{tabular}{lccccccccc}
\toprule
 & \multicolumn{3}{c}{Qwen-Plus (ICC 0.76)} & \multicolumn{3}{c}{Gemini 3 Flash (ICC 0.59)} & \multicolumn{3}{c}{Qwen-Turbo (ICC 0.96)} \\
\cmidrule(lr){2-4}\cmidrule(lr){5-7}\cmidrule(lr){8-10}
Interval & $\mu$ & $\bar p$ & width & $\mu$ & $\bar p$ & width & $\mu$ & $\bar p$ & width \\
\midrule
(a) pooled Wilson & 66.6 & 100.0 & 8.4 & 71.5 & 99.6 & 6.8 & 60.2 & 100.0 & 8.3 \\
(b) two-level boot. & 95.6 & 100.0 & 17.3 & 94.9 & 100.0 & 13.0 & 95.2 & 100.0 & 18.4 \\
(b$'$) one-level & 94.7 & 100.0 & 16.9 & 93.5 & 100.0 & 12.4 & 94.8 & 100.0 & 18.3 \\
(b$''$) clust.\ normal & 94.7 & 100.0 & 17.1 & 92.7 & 100.0 & 12.5 & 94.9 & 100.0 & 18.5 \\
(c) fixed, normal & 34.7 & 93.6 & 4.1 & 48.4 & 94.5 & 4.4 & 11.7 & 91.3 & 1.6 \\
(c$'$) fixed, runs & 31.8 & 89.7 & 3.6 & 45.5 & 90.5 & 3.9 & 10.8 & 86.3 & 1.4 \\
\bottomrule
\end{tabular}
\end{table}

%% file: tables/tab_repeats_disjoint.tex
\begin{table}[htbp]
\centering
\small
\setlength{\tabcolsep}{3pt}
\caption{Identification from $R'$ runs with disjoint estimation and evaluation (\textsc{Detailed}, 14 clean systems, 200 random splits per model). Components are estimated from $R'$ random runs on a random 50-problem half and compared with reference values from the other 49 problems using the $5-R'$ unused runs (rem.; for $R'=4$, all five runs, since one run cannot identify $\sigma_b^2$) or using all five runs (all 5; still independent of the estimate because the problems are disjoint). The $R'=5$ row uses all five runs on both halves and so measures problem-sampling noise between halves alone. MAE excludes splits with $\widehat{\sigma_b^2}\le0$. CI width: median over models of the 95\% problem-bootstrap interval on the ceiling from an $R'$-run pilot on all 99 problems (median over run subsets). PSR error: beta-binomial prediction of five-run retry-free coverage on the held-out half, from $(\hat\mu,\widehat{\sigma_b^2})$ of the estimation half (full) or from its dispersion $\alpha+\beta$ combined with the held-out score (disp.). For scale, the observed five-run PSR of the estimation half misses the held-out PSR by 7.64\,pp: most of the error is problem-sampling noise between halves.}
\label{tab:repeats-disjoint}
\begin{tabular}{lccccc}
\toprule
$R'$ & Ceiling MAE & ICC MAE & $\widehat{\sigma_b^2}\le0$ (\%) & Ceiling CI width & PSR MAE (pp) \\
 & ref.\ rem.\ / all 5 & ref.\ rem.\ / all 5 & & & full / disp. \\
\midrule
2 & 0.272 / 0.249 & 0.108 / 0.101 & 0.0 & 0.54 & 8.28 / 4.58 \\
3 & 0.250 / 0.181 & 0.107 / 0.086 & 0.0 & 0.39 & 7.78 / 3.93 \\
4 & 0.170 / 0.170 & 0.081 / 0.081 & 0.0 & 0.36 & 7.91 / 3.75 \\
\midrule
5 (floor) & 0.160 & 0.077 & 0.0 & 0.33 & 7.47 / 3.63 \\
\bottomrule
\end{tabular}
\end{table}

%% file: tables/tab_ranks.tex
\begin{table}[htbp]
\centering
\small
\caption{Rank uncertainty under the paired two-level bootstrap (\textsc{Detailed}, RLPR, $B=2000$; problems resampled once per replicate for all models, runs resampled within problem). Rank interval: 2.5th--97.5th percentile of the bootstrap rank (ties share the better rank). 3 of 15 adjacent pairs are resolved at 95\%: Gemini 3 Flash $>$ QwQ-Plus ($P=0.989$); DeepSeek-V3.2 $>$ Claude Haiku 4.5 ($P=0.979$); Claude Haiku 4.5 $>$ Qwen-Turbo ($P=0.983$). Under the one-level problem bootstrap 3 or 4 of 15 are resolved, depending on the seed.}
\label{tab:ranks}
\begin{tabular}{lccccc}
\toprule
Model & RLPR (\%) & Rank & 95\% rank interval & $P(\text{top-1})$ & $P(\text{top-3})$ \\
\midrule
Gemini 3.1 Pro & 86.7 & 1 & [1, 3] & 0.57 & 0.98 \\
o4-mini$^{\dagger\ddagger}$ & 86.1 & 2 & [1, 3] & 0.35 & 0.99 \\
DeepSeek-R1$^{\dagger}$ & 84.0 & 3 & [1, 4] & 0.11 & 0.83 \\
Gemini 3 Flash & 81.2 & 4 & [2, 4] & 0.01 & 0.21 \\
QwQ-Plus$^{\dagger}$ & 72.9 & 5 & [5, 7] & 0.00 & 0.00 \\
Gemini 2.5 Flash+Think$^{\dagger\ddagger}$ & 72.3 & 6 & [5, 7] & 0.00 & 0.00 \\
Gemini 3.1 Flash-Lite & 69.1 & 7 & [5, 8] & 0.00 & 0.00 \\
Qwen-Plus & 63.6 & 8 & [7, 10] & 0.00 & 0.00 \\
GPT-4.1 & 59.6 & 9 & [8, 11] & 0.00 & 0.00 \\
GPT-4.1-mini & 59.0 & 10 & [8, 11] & 0.00 & 0.00 \\
Gemini 2.5 Flash & 54.1 & 11 & [10, 12] & 0.00 & 0.00 \\
Claude Sonnet 4.5 & 53.3 & 12 & [10, 13] & 0.00 & 0.00 \\
DeepSeek-V3.2 & 48.7 & 13 & [12, 13] & 0.00 & 0.00 \\
Claude Haiku 4.5 & 41.6 & 14 & [14, 14] & 0.00 & 0.00 \\
Qwen-Turbo & 34.7 & 15 & [15, 16] & 0.00 & 0.00 \\
Qwen-Max & 30.9 & 16 & [15, 16] & 0.00 & 0.00 \\
\bottomrule
\end{tabular}
\end{table}

%% file: tables/tab_independence.tex
\begin{table}[htbp]
\centering
\small
\setlength{\tabcolsep}{4pt}
\caption{Run-label diagnostics (14 clean systems). The run index is a request label within one batch job, not a generation order (\S\ref{sec:setup}). Label effects: Cochran's $Q$ per model (problems as blocks, 4 df), summed across models. Index-adjacent agreement: among problems with $0<\hat p_i<1$, agreement of runs with adjacent indices $(r,r{+}1)$ vs.\ non-adjacent pairs; two-sided permutation test permuting the index within problem (10{,}000 permutations).}
\label{tab:independence}
\begin{tabular}{llcc}
\toprule
\multicolumn{4}{l}{\emph{Label effects}} \\
Prompt & Pass rate by run index 1--5 (\%) & Max gap & $\sum Q$ (df $=56$), $p$ \\
\midrule
\textsc{Detailed} & 60.3 / 59.7 / 60.3 / 60.0 / 59.5 & 0.8\,pp & 42.7, $p=0.90$ \\
\textsc{Minimal}  & 58.4 / 59.5 / 58.1 / 58.7 / 58.8 & 1.4\,pp & 53.8, $p=0.56$ \\
\multicolumn{4}{l}{\footnotesize No individual model has $p<0.05$ (smallest $p=0.079$ \textsc{Detailed}, $0.095$ \textsc{Minimal}).} \\
\addlinespace
\multicolumn{4}{l}{\emph{Index-adjacent agreement}} \\
Prompt & Agreement: adjacent vs.\ non-adjacent runs & Mixed problems & Permutation $p$ \\
\midrule
\textsc{Detailed} & 49.2\% vs.\ 52.0\% & $n=279$ & 0.12 \\
\textsc{Minimal}  & 50.2\% vs.\ 51.7\% & $n=258$ & 0.43 \\
Both              & 49.7\% vs.\ 51.9\% & $n=537$ & 0.09 \\
\bottomrule
\end{tabular}
\end{table}

%% file: tables/tab_ceiling_subsample.tex
\begin{table}[htbp]
\centering
\small
\caption{Stability of the median ceiling (14 clean systems, \textsc{Detailed}) under random problem subsets of size $n$ drawn without replacement (500 draws). Per-model 5--95\% range at $n=50$: Qwen-Turbo 1.02--1.09 (full 1.05); Gemini 3 Flash 1.44--2.24 (full 1.71).}
\label{tab:ceiling-subsample}
\begin{tabular}{lccc}
\toprule
$n$ & 5th pct. & Median & 95th pct. \\
\midrule
30 & 1.20 & 1.31 & 1.45 \\
50 & 1.25 & 1.33 & 1.44 \\
70 & 1.28 & 1.34 & 1.40 \\
99 (full) & --- & 1.34 & --- \\
\bottomrule
\end{tabular}
\end{table}